\documentclass[journal]{IEEEtran}
\ifCLASSINFOpdf
  \usepackage[pdftex]{graphicx}
\else
\fi

\usepackage{csquotes}

\usepackage{color}

\begin{document}
%
% paper title
% Titles are generally capitalized except for words such as a, an, and, as,
% at, but, by, for, in, nor, of, on, or, the, to and up, which are usually
% not capitalized unless they are the first or last word of the title.
% Linebreaks \\ can be used within to get better formatting as desired.
% Do not put math or special symbols in the title.
\title{Scalable Co-Optimization of Morphology and Control in Embodied Machines }
%
%
% author names and IEEE memberships
% note positions of commas and nonbreaking spaces ( ~ ) LaTeX will not break
% a structure at a ~ so this keeps an author's name from being broken across
% two lines.
% use \thanks{} to gain access to the first footnote area
% a separate \thanks must be used for each paragraph as LaTeX2e's \thanks
% was not built to handle multiple paragraphs
%

\author{Nick~Cheney$^{1,2*}$,
        Josh~Bongard$^{3}$,
        Vytas~SunSpiral$^{4}$,
        and~Hod~Lipson$^{5}$
     
$^{1}$ Department of Computational Biology and Biological Statistics, Cornell University, Ithaca, NY

$^{2}$ Department of Computer Science, University of Wyoming, Laramie, WY

$^{3}$ Department of Computer Science, University of Vermont, Burlington, VT

$^{4}$ Intelligent Robotics Group, Intelligent Systems Division, NASA Ames/SGT Inc., Mountain View, CA

$^{5}$ Department of Mechanical Engineering, Columbia University, New York, NY
%\thanks{Manuscript received May 20, 2017; revised ?.}
\\$^{*}$nac93@cornell.edu
}

\maketitle

% As a general rule, do not put math, special symbols or citations
% in the abstract or keywords.
\begin{abstract}

Evolution sculpts both the body plans and nervous systems of agents together over time. In contrast, in AI and robotics, a robot's body plan is usually designed by hand, and control policies are then optimized for that fixed design.  The task of simultaneously co-optimizing the morphology and controller of an embodied robot has remained a challenge.  In psychology, the theory of embodied cognition posits that behavior arises from a close coupling between body plan and sensorimotor control, which suggests why co-optimizing these two subsystems is so difficult: most evolutionary changes to morphology tend to adversely impact sensorimotor control, leading to an overall decrease in behavioral performance. Here, we further examine this hypothesis and demonstrate a technique for \textquote{morphological innovation protection}, which temporarily reduces selection pressure on recently morphologically-changed individuals, thus enabling evolution some time to \textquote{readapt} to the new morphology with subsequent control policy mutations. We show the potential for this method to avoid local optima and converge to similar highly fit morphologies across widely varying initial conditions, while sustaining fitness improvements further into optimization.  While this technique is admittedly only the first of many steps that must be taken to achieve scalable optimization of embodied machines, we hope that theoretical insight into the cause of evolutionary stagnation in current methods will help to enable the automation of robot design and behavioral training -- while simultaneously providing a testbed to investigate the theory of embodied cognition.

\end{abstract}

%% Note that keywords are not normally used for peerreview papers.
%\begin{IEEEkeywords}
%Embodied Cognition, Evolutionary Robotics, Brain-Body Co-Optimization, Soft Robotics, Morphology.
%\end{IEEEkeywords}

% For peer review papers, you can put extra information on the cover
% page as needed:
% \ifCLASSOPTIONpeerreview
% \begin{center} \bfseries EDICS Category: 3-BBND \end{center}
% \fi
%
% For peerreview papers, this IEEEtran command inserts a page break and
% creates the second title. It will be ignored for other modes.
\IEEEpeerreviewmaketitle

\section{Introduction}

%{\color{red} [Nick, your introduction is one long monolithic block of text. Advertise the structure of the intro by breaking into several subsections. I've given some below, but if thesedon't cleave your argument into appropriate pieces, please reorganize as you see fit.]}
%{\color{red} [Remove all mention of evolution, and evolutionary algorithms, as `optimization'. Evolution does not optimize; it satisfices. So, use words like `evolutionary algorithm' instead of `optimization algorithm' and 'evolution' instead of 'biological optimization'.]}

% \edit{***BIOLOGICAL SIGNIFICANCE DISCLAIMER -- doesn't directly model biological phenomenon -- but indirectly related in the following ways}

Designing agile, autonomous machines has been a long-standing challenge in the field of robotics~\cite{pfeifer2007self}.  Animals, including humans, have served as examples of inspiration for many researchers, who meticulously and painstakingly attempt to reverse engineer the biological organisms that navigate even the most dynamic, rugged, and unpredictable  environments with relative ease~\cite{chestnutt2005footstep,raibert2008bigdog,wood2008first}.  However, another competing approach is the use of  evolutionary algorithms to  search for robotic designs and behaviors without presupposing what those designs and behaviors may be. These methods often take inspiration from the evolutionary method itself, rather than the exact specifications of any given organism produced by it.

The use of an evolutionary algorithm for automated design comes with many benefits: 
%{\color{red}Firstly,}
It removes the costly endeavour of determining which traits of a given organism are specific to its biological niche, and which are useful design features that can provide the same beneficial functions if instantiated in a machine.  It can yield machines that do not resemble any animals currently found on earth~\cite{langton1989artificial},
% and is capable of online adaptation -- allowing for the continued optimization and specialization which comes from a robot's own experiences and use cases (which may differ from those of the model animal) 
% {\color{red} [Nick, I'm confused by this last point. Are you saying that evolved robots may be able to adapt over their life times better than animals? Clarify]}.  
as it allows for machines that are specialized for behaviors and environments that differ from those of the model organism.  
Additionally, the optimization process can serve as a controlled and repeatable test-bed for the study of evolutionary, developmental, or behavioral theory~\cite{lenski1999genome,lenski2003evolutionary,bongard2011morphological}.  

%However the approach of bio-inspired optimization also presents challenges that have yet to be overcome, causing
%, thus it tends to scale poorly:
%the scale and complexity of evolved robots and virtual agents to pale in comparison to their biological counterparts. It is unclear what aspects of (or omissions from) current evolutionary algorithms are preventing robotic optimization to scale the way that biological evolution has demonstrated.  

%\edit{Josh -- Q1: should we even talk about problems with the current state of Evolutionary Robotics that we are not going to address in this paper? A: No.}

However, the generalization of design automation to include both the optimization of robot neural controllers and body plans has proven to be problematic.  While recent successes have demonstrated the potential of effective optimization for the control policies of agents with fixed morphologies~\cite{geijtenbeek2013flexible,cully2015robots,levine2016end} or -- to a lesser extent -- the optimization of morphologies (body plans) for agents with minimal and fixed control policies~\cite{collins2005efficient,cheney2013unshackling,auerbach2014environmental}, the co-optimization of the two has seen very limited success~\cite{geijtenbeek2012interactive, cheney15difficulty}.

This inability to perform robot brain-body co-optimization at scale has been experienced and noted informally by many researchers involved with robot optimization, yet published rarely.  Thus the lack of publication is presumably because the field lacks incentives for the publication of negative results~\cite{
%dickersin1990existence,
fanelli2011negative}, rather than a lack of negative results in unpublished works.  Joachimczak et al. provide an anecdotal example of premature convergence in the co-optimization of robot brain and body plan (Fig.~19 of~\cite{joachimczak2016artificial}).  Cheney et al.~\cite{cheney15difficulty} further analyze the phenomenon of premature convergence in embodied machines and suggest that traditional evolutionary algorithms are hindered in this setting primarily in their ability to perform continued optimization on the morphology of the robot.  They hypothesized that the premature convergence may be due to an effect of embodied cognition, in which an individual's body plan and brain have an incentive to specialize their behaviors to complement one another.  This specialization makes improvements to either subsystem difficult without complementary changes in the other (a highly unlikely event given current algorithms) and thus results in an embodied agent which is fragile with respect to design perturbations. 
\section{Background}

Attempts to solve this problem of fully-automated robot evolution are frequently traced back to the work of Sims~\cite{sims1994evolving}.  This work introduced the use of evolutionary algorithms to produce goal-directed behaviors and morphologies simultaneously.  Despite the advance this work represented, the evolved robots tended to be composed of a small number of components (figures show a mean of 6.042 segments per robot, with each typically controlled by a few neurons).  It has been suggested that 
%in the decades since~\cite{sims1994evolving}, increases in 
current computational power
% following Moore's Law~\cite{schaller1997moore} 
should have vastly increased the scale and complexity of robots evolved using Sims' and similar methods.
%~\cite{cheney2013unshackling}.  
Yet 
%vast increases in scale have not been the case empirically, as 
surveys of evolved robots (e.g.~\cite{geijtenbeek2012interactive, bongard2013evolutionary}) fail to exhibit any significant increase in size or complexity.
%~\cite{sims1994evolving,sims1994evolvingCompetition}.  

A wide range of hypotheses for the lack of scalability have been proposed.  Some focused on a lack of efficient evolutionary search algorithms~\cite{hornby2006alps,lehman2011evolving,mouret2015illuminating} or genetic encodings~\cite{hornby2001body,bongard2003evolving,cheney2013unshackling}, while others pointed to a lack of incentives for complexity in the simple tasks and environments~\cite{auerbach2014environmental,cheney2015evolving}.  Yet attempts to evolve robots using methods designed to overcome these challenges have yet to obviously surpass Sims' work in terms of complexity and scale.  This work investigates a different hypotheses, first suggested in~\cite{cheney15difficulty}, that considers the way in which an agent's brain and body plan interact during the optimization process.  

%{ \color{red} [remove paragraph for space?] }
%This work is related to co-evolutionary methods (such as~\cite{koza1991genetic, sims1994evolvingCompetition, cliff1995tracking, dieckmann1995evolutionary, keerativuttitumrong2002multi, omidvar2014cooperative}) in that there are multiple components being evolved.  These classical co-evolutionary systems (like predator-prey) typically call for two separate populations being evolved as generalists against one another.  However, for the case of brain-body co-optimization in evolutionary robotics, it is often desirable to have the body and brain of a robot specialized to one another, and developed simultaneously from a single genome, and in a single population (as in~\cite{sims1994evolving, lipson2000automatic, hornby2001body, auerbach2010dynamic, bongard2003evolving} -- and as is the case in natural systems).  This combined brain-body genome causes particular optimization issues which we hypothesize below.  

\section{Methods}

\subsection{The Interdependency of Body and Brain}

%Specifically, we
We investigate the notion that the specialization of brain and body plan to one another during evolution creates a fragile co-dependent system that is not easily amenable to change.  This specialization creates local optima in the search space and premature convergence to suboptimal designs.  In this paper, we explore a direct solution to the problem of fragile coupled systems: explicitly readapting one subsystem (e.g. the body plan or the brain) after each evolutionary perturbation to the other.  The proposed method differs from a traditional evolutionary algorithm, which evaluates the fitness of a newly proposed variation immediately (i.e. with no readaptation), and uses only this valuation of fitness to determine the long term potential of that variation.  

As a thought experiment, consider a hypothetical robot with a partially optimized body plan (for example a quadrupedal form) 
and a partially optimized controller (for example the legs swing forward and back through the sagittal plane).  Suppose that this controller has co-adapted to the morphology during evolution such that for each step forward, the robot contracts a muscle near its hip with enough force to swing one of its legs forward with just enough force to land in front of the body and successfully take  a step.  Consider a variation of this morphology proposed by an evolutionary algorithm in which the new robot possesses longer legs but the controller has not changed.

If this machine were being evolved for rapid locomotion, having longer legs and being able to take longer strides would be beneficial and this variation should be more successful than the original design. However, during evaluation of this new robot, the original controller applies the same amount of force to the now-longer leg, failing to move it, and thus frustrating the robot's ability to walk in a coordinated fashion.  Current evolutionary methods would treat this robot as the recipient of a detrimental mutation and remove it from the population.  
%
%In this example, a mutation to one subsystem (the body plan) could be beneficial to the robot's descendants, but since the immediate impact of the change is detrimental, the mutated robot suffers a decrease in fitness and does not survive to produce offspring. 
If all such variations display an immediate detrimental fitness impact and are rejected (regardless of their long term potential), then the evolutionary algorithm has 
%prematurely 
converged to a local optima in the search space.
% because there appear to be no better alternatives in the immediate neighborhood of the current design.  

However, the newly proposed morphology would have resulted in a robot which outperformed its predecessor, if coupled with a controller suited to that morphology.  We can determine that the newly proposed morphology is superior by suppressing mutations to that body plan and allowing readaptation of its controller to properly coordinate behavior.
%{\color{red}\st{Upon performing optimization on only the controller for a given amount of time, we would see that the robot again has learned to contract its muscles with enough force to swing its (now longer) leg all the way forward with each step -- resulting in longer strides and thus faster locomotion (assuming a constant oscillation frequency).}}
%over phylogenetic time.

Herein lies the foundation for our proposed algorithm: readapt each controller until the new proposed morphology is more fit than its predecessor.

%\section{Methods}

\subsection{Controller Readaptation}

% {\color{red}
% \st{But what happens if the proposed morphology had shortened each of the legs, or shortened some while lengthening others?  Regardless of the amount of control readapatation, the fitness of the child would never outperform its parent, as the mutation was indeed a detrimental one.  By evaluating the success of each proposed morphology after a given amount of control readaptation (x iterations of control optimization), we are able to fairly access the long term potential of how effective that morphology could be with a properly adapted controller -- where the threshold for long term is defined as x iterations of optimization.}}
The most obvious method for modeling controller readaptation would be to protect any lineage that has recently experienced a mutation to the body plan by allowing it to undergo several generations of evolutionary change restricted to the control subsystem. If any member of the lineage achieves higher fitness than the pre-mutation ancestor during that time period, the descendant is retained. Otherwise, the new morphological variant dies out.

% {\color{red}
% \st{But how long should we set this threshold of time to search for controllers after each new morphological mutation?}}
However, it is unclear how to set the time period for this protection \textit{a priori}. Surely the amount of time a controller takes to readapt to a new morphology depends on many specific features of the complexity, genetic encoding, desired behavior, and current ability level of the robot (which changes over optimization time).  Determining the correct value of this parameter would require a full parameter sweep over various values of readaptation time for each new combination of brain, body, and environment.  If our goal is simply to optimize a robot, then searching for this value in each unique optimization scenario is intractable. 

%\subsubsection{\color{red} The proposed method.}
\subsection{Proposed Method: Morphological Innovation Protection}

% {\color{red}
% \st{Instead we will allow this parameter to vary.  Rather than readapting a controller for a fixed amount of optimization each time a newly proposed morphology is presented and then making a single comparison to its parent, we will continually introduce new potential morphologies into a population of partially optimized robots, and perform continual comparison between these different morphologies as they readapt their controller -- but making sure to consider a morphology to be outcompeted (and thus remove it from the population) only if there is a more effective new morphology that has undergone less control readaptation than itself (which is implemented as a multi-dimensional optimization problem}
% Because of this, we instead propose a parameter-free approach... [Wait, I just realized/remembered
% that your approach is not parameter-free. Nick, you're going to have to include some
% results here showing that your parameter (the percentage of body mutation that is
% worthy of control readaptation) is less sensitive than the parameter in the naive
% method (the number of fixed generations alloted for control readaptation.)]}

%\edit{add more details/context to this paragraph -- it's the first time our method is introduced!}
In response to the unintuitive nature of the optimal value for readaptation length, 
our proposed approach is free of this parameter.   
Descendants of robots that experience morphological mutations are retained in the population and the number of generations that have elapsed since that mutation occurred are tracked (referred to as the \textquote{morphological age} of the robot). If two individuals are found in the population such that the latter robot exhibits better performance on the desired task and has experienced fewer generations since a morphological mutation than the former robot, this latter robot is said to \textquote{dominate} the former robot in this mutli-objective optimization (on \textquote{age} and \textquote{fitness}) -- and the former robot is removed from the population. In effect, the latter robot has exhibited an ability to not only recover from its ancestor's morphological mutation but improve upon it (and others in the population).  The concept of tracking age using it as an optimization objective are borrowed from~\cite{hornby2006alps, schmidt2011age}.  The major difference here being that the age refers to the length of time that a subsystem of the agent (e.g. the morphology) has remained unvaried, rather than (the original definition of) the total time since a random individual was introduced to the population.   
%More methodological details are provided in \textquote{Morphological Innovation Protection}).
%See \textquote{Methods} for more details.    

This procedure has the effect of \textquote{protecting} new morphologies with poorly adapted controllers, and thus we will henceforth refer to this procedure as \textquote{morphological innovation protection.}  This protection is a form a diversity maintenance, though reduced selection pressure for newly mutated morphologies.  Various other methods exist for encouraging diversity (e.g. fitness sharing, crowding, random restart parallel hillclimbers~\cite{gordon1993serial}, novelty~\cite{lehman2011evolving}, or speciation~\cite{stanley2002evolving}), however age was chosen for its simplicity of implementation, its parallels to multi-timescale learning in biology, and because it helps to avoid the cost of extended control re-optimization for non-promising morphologies -- since the the age-pareto optimization allows fitness comparisons between all new \textquote{child} morphologies that have had equal readaptation time, even if they are not yet fully readapted (rather than making comparisons only after a predefined amount of readaptation).  

%{ \color{red} [TODO: talk about relation to ALPS/AFPO] }

\subsection{Evolutionary Algorithm}
All optimization is performed by a population-based evolutionary algorithm. 
%, inspired by the popular algorithm NEAT (NeuroEvolution of Augmenting Topologies)~\cite{stanley2002evolving} (importantly without speciation).  
All trials follow a $(\mu,\lambda)$-Evolutionary Strategy~\cite{beyer2002evolution} with $\mu=25$ parents and $\lambda=25$ mutants for a population size of 50.  Trials last for 
% 1000 generations in the simplier case of phase-offset controllers and 
5000 generations. 
% for the more complex case of neural network controllers (see \textquote{Phase Offset Controllers} and \textquote{Neural Network Controllers} below).  
Crossover was not considered in this work.  Mutation had a 50\% chance of creating a variation to either the morphology or the controller of a given robot, but not both.  Other ratios of morphology:controller mutations were considered (1:99, 20:80, 50:50, 80:20, and 99:1), but none showed a significant effect on resolving the premature convergence and resulting fitness in preliminary trials without innovation protection.  

\subsection{Genetic Encoding for Soft Robot Morphologies}
Consistent with prior work studying the co-optimization of robot morphologies~\cite{cheney2013unshackling} and controllers~\cite{cheney15difficulty}, we choose soft robots as our model system due to the open-ended complexity of deformable voxel-based morphologies and distributed controllers.  The soft robot morphologies are encoded with a Compositional Pattern Producing Network (CPPN)~\cite{stanley2007compositional}.  The CPPN encoding produces the cell fate of each voxel in the robot through a type of neural network that takes each cell's geometric location ($x, y, z$ Cartesian coordinates and $r$ radial polar coordinate) and outputs a variety of \textquote{morphogens} (in this work, there is one to determine whether a cell is present in that location and one to determine whether a present voxel should be a muscle or a passive tissue cell).  Since nearby voxels tend to have similar coordinate inputs, they also tend to produce similar outputs from the network -- creating continuous muscle or tissue patches.  CPPNs also produce complex geometric patterns, as the activation functions at each node can take on a variety of functions (here: \textit{sigmoid, sine, absolute value, negative absolute value, square, negative square, square root}, and \textit{negative square root}).  These functions tend to produce regular patterns and features across the coordinate inputs (for example: an \textit{absolute value} node with an $x$ input would produce left-right symmetry, or a \textit{sine} node with a $y$ input would produce front-to-back repetition).  

This network is optimized to produce high performing morphologies by iterating through various proposed perturbations to it.  These include the addition or removal of a node, or edge to the network, as well as the mutation of the weight of any edge or the activation function at each node.  

% repeated below, so cut for space
%As the CPPN is an indirect and generative encoding, the genotype (neural network) is independent of the scale of the phenotype (soft robot being built), since the network is a continuous function that is discretized only to be queried once per voxel.
%, regardless of the number of voxels.  
%This means that the optimized network is scale free, and can be applied to build robots with any number of voxels (i.e. any resolution or overall size) -- which is limited only by available physical resources (for real world robots) or computation resources (in the case of simulated robots here).  

\subsection{Soft Robot Resolution}
As the CPPN genetic encoding is a continuous function (mapping the location of a cell to its cell fate) it may be discretized into a phenotype at any resolution (i.e. creating any number of voxels in the morphology, and a unique controller for each voxel), and in practice this resolution is only limited by computational resources (as more elements are more computationally expensive to simulate).  In the
default lower-resolution treatment,
%longer-running treatments (Figs.~\ref{fitnessOverTime}-\ref{raindowWaterfallControl}) 
this discretization occurs over a $5 \times 5 \times 5$ space.  The higher-resolution robots
%, and for more detailed viewing of phenotypes, Figs.~\ref{morphologiesOverTimeWithProtection}-\ref{morphologiesOverTimeNoProtection} 
use phenotypes created at a $10 \times 10 \times 10$ resolution.  Note that the distance values are noted in absolute number of voxels, 
%and voxels are held at a constant size regardless of resolution -- one centimeter in our simulations, 
so higher fitness values tend to be produced by phenotypes of higher resolution.
%, as more muscle mass is available to those individuals.  

\subsection{Controllers and their Genetic Encoding}
%In order to control the robots (determine when to contract each muscle cell -- and with what speed and force), two different controller strategies were considered for different experimental treatments:  a simpler open-loop oscillatory controller, and a more complex closed-loop neural network controller.  For each of these control strategies, 

A unique controller is optimized for each muscle cell in the robot's morphology.  
%However, since the parameters of the controller at every voxel of the robot are optimized by a single CPPN (a separate CPPN from the one which produces the robot morphology) -- which is a scale free encoding, the number of unique controllers being optimized is not substantially more costly than optimizing a single global controller.
%
For each voxel's controller, two parameters are optimized (the outputs of a separate CPPN with the same inputs as before).  One is the phase offset between each individual cell's muscle oscillations and that of the global sinusoidal oscillator (which acts as a central pattern generator), and the second is the frequency of this global clock (since CPPNs don't currently enable global parameters, this is done by averaging the local values at each cell to produce a single global value).  

% (which would have a variety of inputs, depending on the current morphology).  
 All controllers output a value between $-1$ and $1$ at each time step, which corresponds to a linear change in each dimension of a muscle cell ($\pm 14\%$ of its original length, or $\pm 48\%$ of its original volume), defining a robot's behavior.  Passive tissue cells remain at their original size (though they also deform based on their intrinsic compliance).

While this encoding is simple and straightforward, it has the ability to produce complex behaviors, such as multiple patches of muscle groups that are in sync, counter-sync, or any real valued phase offset from each other.  It also has the ability to produce gradually varying sweeps of phase offset, resulting in propagating waves of excitation across large muscle groups.  Furthermore, the optimization of the global frequency is able to produce oscillation speeds which are fine tuned to the properties of individual morphologies (such as optimizing to maximize the resonance of soft tissues in appendages).  

The encoding of the morphology and the controller of the robot into two separate CPPN networks emphasizes the false dichotomy of robot brains and body plans.  However, this explicit separation allows us to make changes specific and isolated to either the brain or the body.  This is necessary for the proposed algorithm, as controller readaptation requires iterating through controller changes without affecting the morphology.

\subsection{Physics Simulation for Evaluation}
Once the morphology and controller for a given robot are specified, the fitness (locomotion distance) of that robot is determined by constructing and simulating that robot in the VoxCad soft-body physics simulator~\cite{hiller2014dynamic}.  Simulations last for 20 actuation cycles (which may be a variable amount of time, depending on the length of the globally optimized frequency -- though this method of normalizing for the number of \textquote{steps} taken leads to a more fair comparison than normalizing by the amount of time per simulation).  

\subsection{Morphological Innovation Protection}

%The method of incorporating \textquote{morphological innovation protection} is closely related to the idea of using \textquote{age} in~\cite{schmidt2011age}, in that it adds a secondary (non-fitness) objective to artificially transform a single-objective optimization into a multi-dimensional search problem.  In prior works, this age objective has been used to maintain diversity in a population as new random genotypes were continually inserted into that population.  By performing Pareto selection, and only considering individuals to be dominated if another individual in the population has both higher fitness and lower age (measured in generations), this emphasizes exploration of the search space by effectively starting a number of \textquote{random resets} and comparing those individuals to the fitness levels that other random starting points would have achieved after similar amounts of optimization.  The advantage of combining these random individuals into a single population rather than performing truly independent parallel random resets is that the comparisons allow the deletion of individuals who are deemed to be less promising in the long term (if they have make less progress than others in the same amount of adaptation time), and the continual injection of new individuals encourages open-ended exploration.  Thus this method allows continued and efficient search over rugged landscapes.  

In our newly proposed method %, rather than injecting completely random genotypes with age zero, 
we set the \textquote{morphological age} to zero for each new \textquote{child} that was the result of a morphological mutation to a current individual in the population. 
%(subject to the threshold noted above).  
This means that for an individual to have a large value in their age objective, that individual must have been the result of a large number of successive controller mutations without any change to that robot's body plan.  This setup thus allows a simple comparison method for individuals who have had similar amounts of controller (re)adaptation to their current morphologies -- as a dominated individual would have to have been out-competed in fitness ability by a morphology that is paired with a controller that is less-well adapted to it. %  As before, t
This diversity maintenance mechanism encourages the exploration of new peaks in the rugged landscape of brain-body plans -- with the implicit assumption that unique morphologies correspond to peaks in this landscape.  

The method of age resets corresponding to morphological mutations to existing individuals differs from the prior technique (e.g.~{\cite{hornby2006alps, schmidt2011age}}) of inserting completely random individuals, as it allows the improvement in the fitness levels of age-zero individuals over time.  In the case of traditional age-fitness optimization, age-zero individuals are drawn from the same distribution of fitness values for random genotypes regardless of when they are created.  However, in the case of morphological innovation protection, age-zero individuals are not random and inherit many of the properties of their parents -- meaning that they show higher fitness values over time.  To demonstrate this empirically, we performed linear regression on the age-zero individuals from Fig.~\ref{rainbowWaterfallProtection} (containing morphological innovation protection), which showed a significant ($p<0.001$) increase in fitness over time (from 18.715 at generation 0 to 22.916 at generation 5000).  This confirms a major difference between our proposed technique and the standard approach of age-based diversity maintenance through the introduction of random individuals.  

%; $r^2 = 0.211$

%%omitted for definition in text
%\subsection{Controller Innovation Protection}
%
%As a control method, we also consider a procedure for \textquote{controller innovation protection}, which is similar to that of morphological innovation protection in all regards, except that the \textquote{protection age} is reset to zero when a child results from a mutation to the controller (rather than a mutation to the morphology).  This means that individuals with large values for the protection age objective are those who have had the same controller for a large amount of time and have thus made many attempts at morphological (re)adaptations to that controller.  

%% desribed later in text
%\subsection{Morphological Change Threshold}
%%{\color{red} [move to results section?]} 
%In treatments with the minimum morphological change parameter, an individual's \textquote{morphological age} is only set to zero if a morphological change occurs that affects at least the proportion of voxels dictated by the threshold.  This proportion is relative to the total possible number of voxels (1000 or 125 in the experiments above), rather than the number of voxels present in any particular robot (to help minimize the number of small and non-functional morphological changes for robots composed of few voxels).  

\subsection{Statistical Analysis}
All treatments were performed for 30 independent trials, with random seeds consistent between treatments.  
%As normality of results cannot be assumed, 
All plots show mean values averaged across the most fit individual of 30 trials for each condition with shaded areas representing 95\% bootstrapped confidence intervals of this average, and all p-values are generated by a Wilcoxon Rank-Sum Test~\cite{wilcoxon1964some}.

% Results and Discussion can be combined.
\section{Results}
% Please do not create a heading level below \subsection. For 3rd level headings, use \paragraph{}.

\subsection{The Effect of Morphological Innovation Protection on Fitness}

For the task of locomotion ability (a standard task in evolutionary robotics~\cite{bongard2013evolutionary}), we 
%optimize robots for the maximum distance they travel with 20 oscillations of their muscles (measured in voxel lengths -- corresponding to one centimeter).  We 
first optimize the robots using the traditional method of \textquote{greedy} fitness evaluations for our selection criteria (where immediate locomotion ability determines survival in the population of candidate morphologies -- i.e. with no \textquote{innovation protection}).  In this setup, the traditional method produces robots with an average fitness of 21.717 voxels (with 95\% bootstrapped confidence (CI) interval of 19.457 to 22.426 voxels).

Additionally, we optimized robots in the same task and environment setup, but this time using \textquote{morphological innovation protection} for our selection method -- in which individuals can only be out-competed by those with equal or lesser amounts of controller (re)adaptation to their current morphologies.  This treatment produces significantly more effective robots ($p=6.067*10^{-6}$), with a mean distance travelled of 31.953 voxels (and 95\% CI of 28.157 to 36.511 voxels).  

The increase in fitness shows that morphological innovation protection is a more effective way of optimizing robots, yet it does not conclusively demonstrate that the intuition of~\cite{cheney15difficulty} is correct, as that work demonstrated the asymmetric difficulty in optimizing the morphology of a robot (as compared to optimizing its controller) and drew the hypotheses that this because the morphology encapsulated the controller (acting as a translator between the \textquote{language} of the \textquote{cognitive} functions and the outside environment).  While the above experiment does help to support the intuition that the controller must readapt to a new morphological encoding,
%~\cite{cheney15difficulty}, 
it also introduces confounding effects, such as the added population diversity afforded by \textquote{protection} and the added dimensionality of the search space from its this protection age -- moving search from a single-objective to multi-objective optimization problem.  

To tease apart the influence of these two confounds, we present a treatment where the controllers of the robot undergo an equivalent protection to which the morphologies did in the above experiment.  In this treatment, individuals can only be out-competed by others whose morphology has had equal or lesser amounts of readaptation to their newly mutated controllers -- deemed \textquote{controller innovation protection}.  This condition provides the potential advantages of multi-dimensional search and the added diversity from temporary reductions in selection pressure.  Yet it does not rely on the idea of a broken \textquote{morphological language.}
% proposed by~\cite{cheney15difficulty}, which suggests the key role of the body as the interface between the brain and the environment (and thus the need to explicitly protect it).  
% Under the condition of \textquote{controller innovation protection}, the 
In this treatment,
robots locomote 22.049 voxels on average (95\% CI of 20.726 to 22.641 voxels), which fails to show a significant improvement over the single-objective case of no protection ($p=0.240$), and performs significantly worse ($p=1.211*10^{-4}$) than the protection of morphological innovations.  

\begin{figure}[t!]
\includegraphics[width=\linewidth]{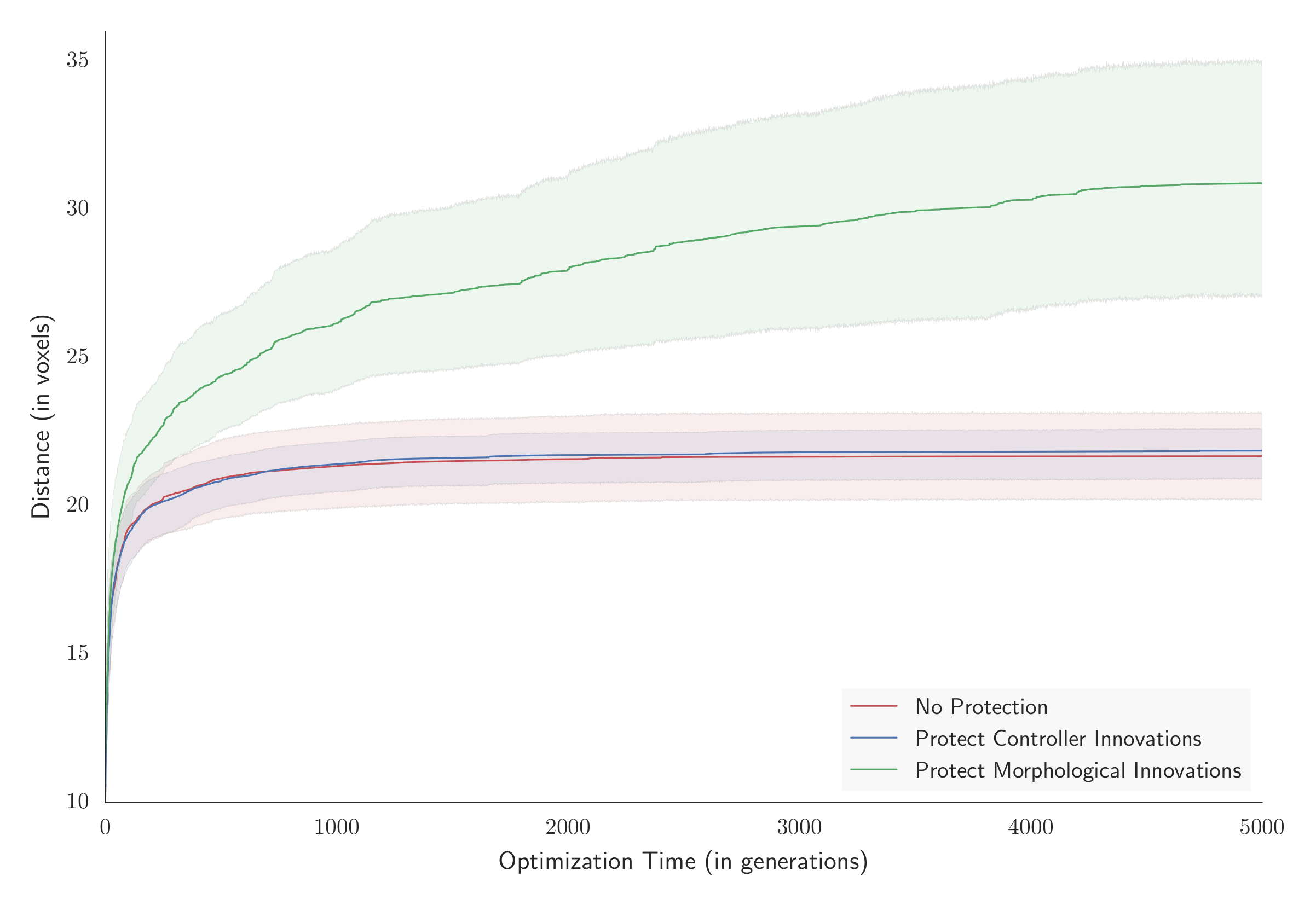}
\caption{The fitness impact (distance travelled, in voxels) over optimization time (in generations) for various types of brain/body plan protection mechanisms.  Values plotted represent the mean value of 30 independent trials, with 95\% bootstrapped confidence intervals denoted by colorized regions.  
%Note that at the end of optimization time (5000 generations), the \textquote{morphological innovation protection} (readapting controllers to new morphologies before evaluating their long term potential) significantly outperform all other treatments ($p < 0.001$), while the \textquote{controller innovation protection} does not perform significantly better ($p=0.240$) than the case with no protection.  This suggests that the added efficacy of protecting morphological innovations goes beyond the effects of an added dimension to the search space, and further suggests an fundamental asymmetry between the morphology and controller while optimizing an embodied robot.
}  

%Also note that protecting controller innovations (by readapting morphologies to each new proposed controller) does not perform significantly better than simply protecting random individuals ($p=0.183$)}.%, suggesting that the significant improvement of controller innovation protection over the baseline case of no protection ($p=0.005$) may be due primarily to the added dimensional of an optimization process with protection than to a careful selection of which individuals do or do not represent long term potential (and thus need to be maintained in the population despite short therm weaknesses).}
\label{fitnessOverTime}
\end{figure}

%Since morphological innovation protection and controller innovation protection both have the added dimension of \textquote{protection age}, the fact that they differ significantly from each other (and that controller innovation protection does not show significant improvements over the baseline case of no innovation protection) suggests that the fitness benefits of the morphological innovation protection are due to more than simply the added dimensionality of the search space and the added population diversity from reduced selection pressure.  Such a result supports the idea of an asymmetry between morphological and control optimization, and demonstrates that the most effective solution to embodied robotic optimization includes the protection of morphological innovations.  

%The full comparison of these fitness values over optimization time are shown in Fig.~\ref{fitnessOverTime}.  
The fitness trajectories over evolutionary time (Fig.~\ref{fitnessOverTime}) demonstrate a typical logarithmic fitness improvement over the first 1000 generations or so, but then show a stagnation for the traditional optimization procedure without innovation protection. %while the fitness values of 
%The treatment with morphological innovation protection contrasts this results by demonstrating sustained improvement further into the optimization period.  
The mean fitness value of the treatment without protection 
shows no significant improvement ($p=0.085$) from generation 1000 to 5000 (with average fitness values of 20.988 and 21.717 voxels, respectively).  
%This finding is consistent with the type of stagnation noted to be traditional in the field of evolutionary robotics in ~\cite{cheney15difficulty}.  
Contrary to this, the treatment with morphological innovation protection shows significant improvements over this time ($p=0.013$) from 25.925 at generation 1000 to 31.953 voxels at generation 5000. 

%Somewhat interestingly, the improvement in the controller innovation protection treatment is also significant ($p=0.017$) over the same period of time, through upon visual inspection this appears to be due to tight confidence interval bands more so than drastic fitness changes as the value improves from a mean of 21.385 to 22.049 voxels.  

The rapid improvement in the controller innovation protection and no protection cases during the first 1000 generations does not contradict the hypotheses of a fragile \textquote{morphological language,} as the coupling between the morphology and controller takes time to become established -- and would not introduce fragility into the system before then.

\subsection{The Effect of Morphological Innovation Protection on Population Stagnation}

%The early stagnation of traditional evolutionary robotics without protection is indicated by the flatline in fitness value in Fig.~\ref{fitnessOverTime}, and suggests the prevalence of local optima in this space.  The treatments which feature innovation protection of some sort appear to stave off stagnation to a greater degree, with visually noticeable results and significantly higher fitness values resulting from trials with morphological innovation protection.  However, these averaged statistics provide little mechanistic insight into why and how protection is able to overcome the local optima. 

\begin{figure}[t!]
\includegraphics[width=\linewidth]{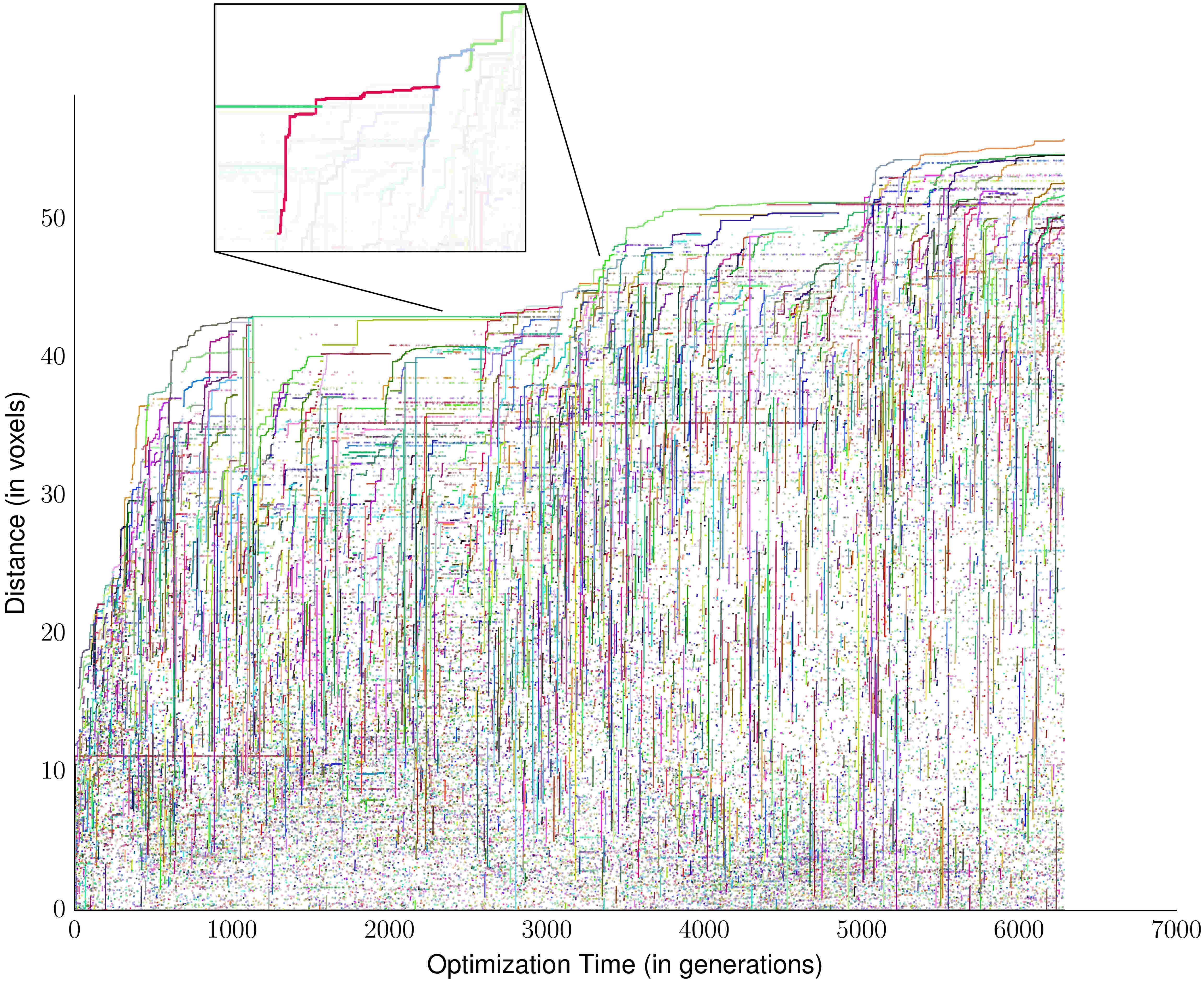}
\caption{A single optimization trial featuring morphological innovation protection.  Each unique morphology is represented by a random color.  
%Note the continual improvement in the locomotion ability over optimization time, and the continual turnover in morphology of the top performing individuals.  
The pop out in this figure highlights an example of an \textquote{overtake} where the new child morphology (in red) initially performs worse than its parent morphology (in teal), only to outperform that previous morphology after successive control optimization to both the parent and child.  
%In traditional greedy methods this initial drop in performance would signify a poor solution and thus remove the child morphology from the population -- leading to stagnation.  The example in the pop-out supports the idea that the continued improvement in the morphological innovation protection treatment is due in part to the ability to properly recognize the long(er) term potential of initially detrimental mutations and allow for such overtake events.
}
\label{rainbowWaterfallProtection}
\end{figure}

\begin{figure}[t!]
\includegraphics[width=\linewidth]{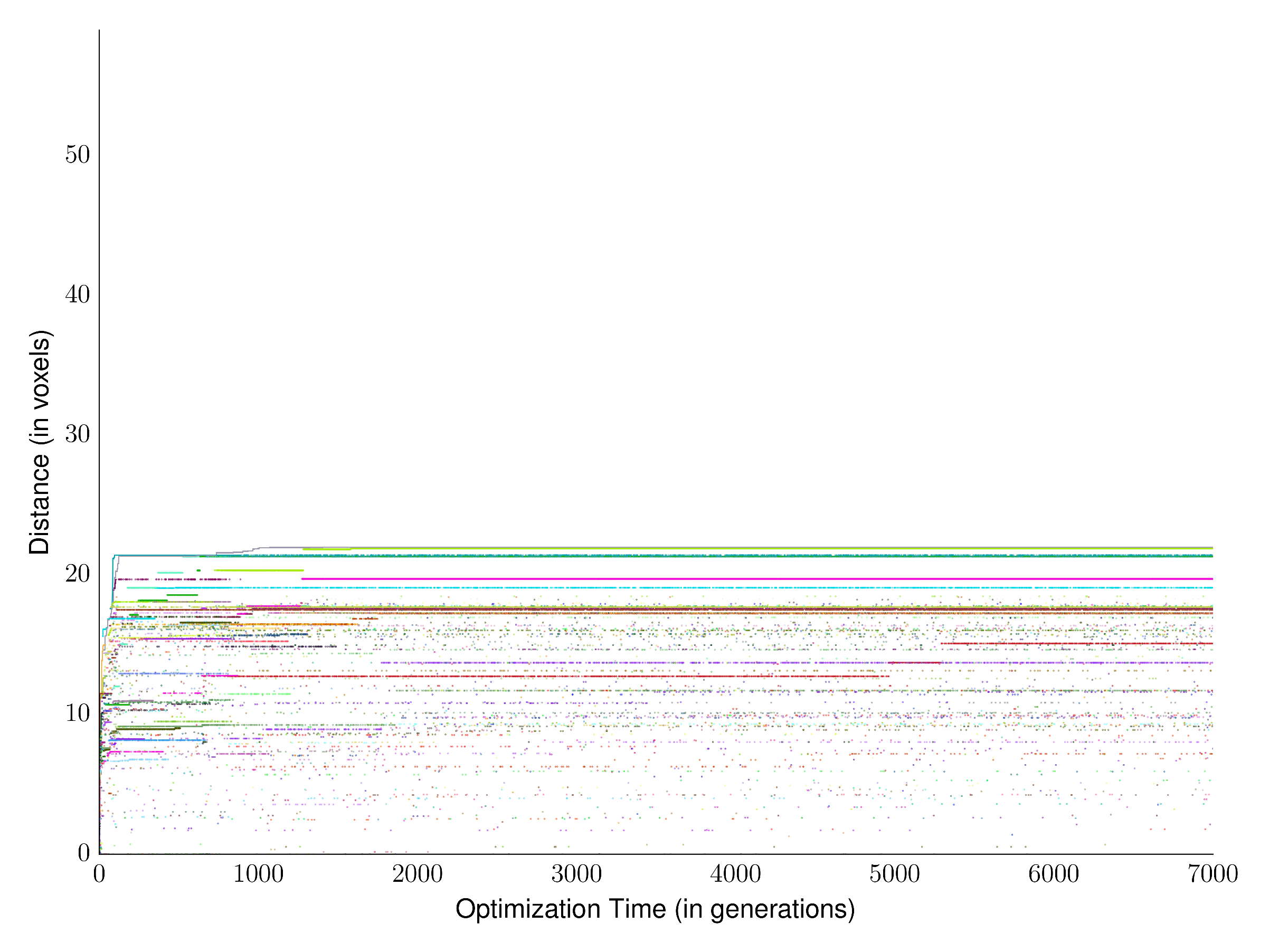}
\caption{The fitness over time of a single optimization trial with the same initial conditions as Fig.~\ref{rainbowWaterfallProtection}, but without any innovation protection.  
Again, each unique color represents a unique morphology.  
%Note the early improvement in locomotion ability during the initialization period (the first couple hundred generations), but the stagnation which occurs immediately afterwards.  Also note 
The prevalence of colored dots filling the space underneath the best performing individual, which represent new morphologies which initially performed worse than the current best individual and were thus rejected and thrown out of the population (as opposed to those individuals which were protected and eventually led to \textquote{overtakes} in Fig.~\ref{rainbowWaterfallProtection}).  
%This treatment -- where morphology and controllers are attempted to be optimized simultaneously and without innovation protection -- represents the currently employed method in the field of evolutionary robotics~\cite{bongard2013evolutionary}.
}
\label{raindowWaterfallControl}
\end{figure}

Perhaps more telling than the average locomotion ability at the end of optimization time is the examination of the optimization process within each individual run.  Figs.~\ref{rainbowWaterfallProtection} and~\ref{raindowWaterfallControl} represent typical runs, and help to give an intuition of the optimization process.  In these figures, each randomly colored line represents a unique morphology, plotted by its locomotion ability over optimization time.  

%Note that these runs outlast the 5000 generation total from Fig~\ref{fitnessOverTime} -- as all trials were optimized for 5 days of walltime, before being truncated for comparisons with other runs.  Some trials reached as far as 10,000 generations, while the minimum number completed was 5014 generations, thus 5000 generations was chosen as the cut off for comparisons of treatment averages -- but figures for single trials show their full length.

%The first thing to n
Note the continued improvement in performance of the most fit individual over optimization time in the case of morphological innovation protection (Fig.~\ref{rainbowWaterfallProtection}) -- which is consistent with the trend seen on average in Fig.~\ref{fitnessOverTime}.  This is not seen in the case without innovation protection (Fig.~\ref{raindowWaterfallControl}), where the best individual was found before generation 2000 -- and by generation 1000, fitness has reached 99.6\% of its final value.  

Consistent with the above observation, we see that the most fit individual in Fig~\ref{rainbowWaterfallProtection} changes rapidly, continually turning over in the trial with morphological innovation protection.  As each color in the figure represents a unique morphology, we notice that a wide variety of different morphologies hold the title of \textquote{best-so-far}.  On average, in runs with morphological innovation protection, 24.179 unique morphologies are the best-so-far at some point in optimization, where the runs without protection show significantly ($p=1.555*10^{-6}$) less turnover of morphologies, with just 10.115 unique robot body plans doing so.  

The question of how temporary reduced selection pressure (via the morphological-age dimension) of morphological innovation protection may help to improve overall fitness and continued optimization may be best demonstrated in the pop out box for Fig.~\ref{rainbowWaterfallProtection}.  Here we see the current best morphology, in teal.  This morphology was unable to improve on itself for some time, as we see its fitness value (\textit{y-xais}) flatlining.  This \textquote{parent} morphology has a \textquote{child,} a new proposed variation of its morphology, highlighted in red.  As the original fitness value of this morphology 
%(its leftmost point, as the \textit{x-axis} represents optimization time) 
falls below its parent, this individual empirically shows worse performance than its parent -- and thus would not be considered as a viable solution in a traditional evolutionary method.  However, since this new morphology does not have a controller that is well adapted to it (as the controller is specialized for the previous morphology, in teal), morphological innovation protection does not expect this new robot to immediately outperform its parent, and keeps this individual in consideration as one which could hold long term potential but does not show immediate promise.  

Indeed, we see that after a number of controller optimization iterations later (occurring in equal amounts to both the parent and child during this intermediate period), the child morphology (in red) overtakes the parent morphology (in teal) -- achieving higher fitness and demonstrating that it did indeed hold a better long term potential than its parent, despite the immediate drop in fitness.  As the fitness of the parent
% (which has had more time to specialize its controller to its morphology) 
is outperformed by the child (which has had less time to fine-tune its controller to its morphology), we assume that the parent is unlikely to be the most promising robot body plan in the long run, and thus remove it from the population.

We see this trend of \textquote{overtakes} -- where children start out with worse fitness than their parents, but eventually outperform them -- continuing throughout this run (as the blue child then overtakes the red parent, and green overtakes blue in the pop out of Fig.~\ref{rainbowWaterfallProtection}).  We see morphological overtakes significantly ($p \leq 6.939*10^{-10}$) more often in runs with  morphological innovation protection (an average of 76.714 overtakes in the first 5000 generations) than without any innovation protection (where there are only 1.432 overtakes) or in the trials with controller innovation protection (where there are just 1.333 morphological overtakes on average).  

Interestingly, the number of morphological overtakes in the morphological innovation protection treatment is not significantly different ($p=0.533$) than the number of controller overtakes in the controller innovation protection treatment (where a \textquote{child} is a robot with a new \emph{controller} variation, that readapts its morphology to catch back up to its \textquote{parent} controller -- which happens 74.542 times on average).  Combined with the finding in Fig.~\ref{fitnessOverTime} that morphological innovation protection outperforms the other two treatments, this suggests a greater potential for the relative importance of morphological overtakes over controller overtakes -- and again reinforces the asymmetry between morphologies and controllers from an optimization perspective.

\subsection{The Effects of Morphological Innovation Protection on the Progression of Morphologies over Evolutionary Time}

%\edit{change to section on non-zero threshold}
%\edit{runs here use phase offset, not neural network}

%In the above example, we saw continued improvement throughout the length of optimization and search covering a larger number of unique morphologies.
% This is presumably possible by allowing search to continually find new and better morphologies (as represented by the turnover of colors in Fig.~\ref{rainbowWaterfallProtection}).  But for this trend to continue presupposes that there are always new and better morphologies to be found in the search space.  
% The ability to continue improvement longer, and search over more unique morphologies 
%This suggests that evolution with morphological innovation protection is better able to escape local optima. % and find more optimal solutions.%, but this is not necessarily evident from the evidence above.  
%Consider a different example where the desired behavior is more specific:  instead of simply moving as fast as possible, the robot must also stand-up at tall as possible (measured by the height of its center of mass) and also use as little energy as possible (measured by the number of actuated muscle cells it uses).  \edit{?}

\begin{figure}[t!]
\includegraphics[width=\linewidth]{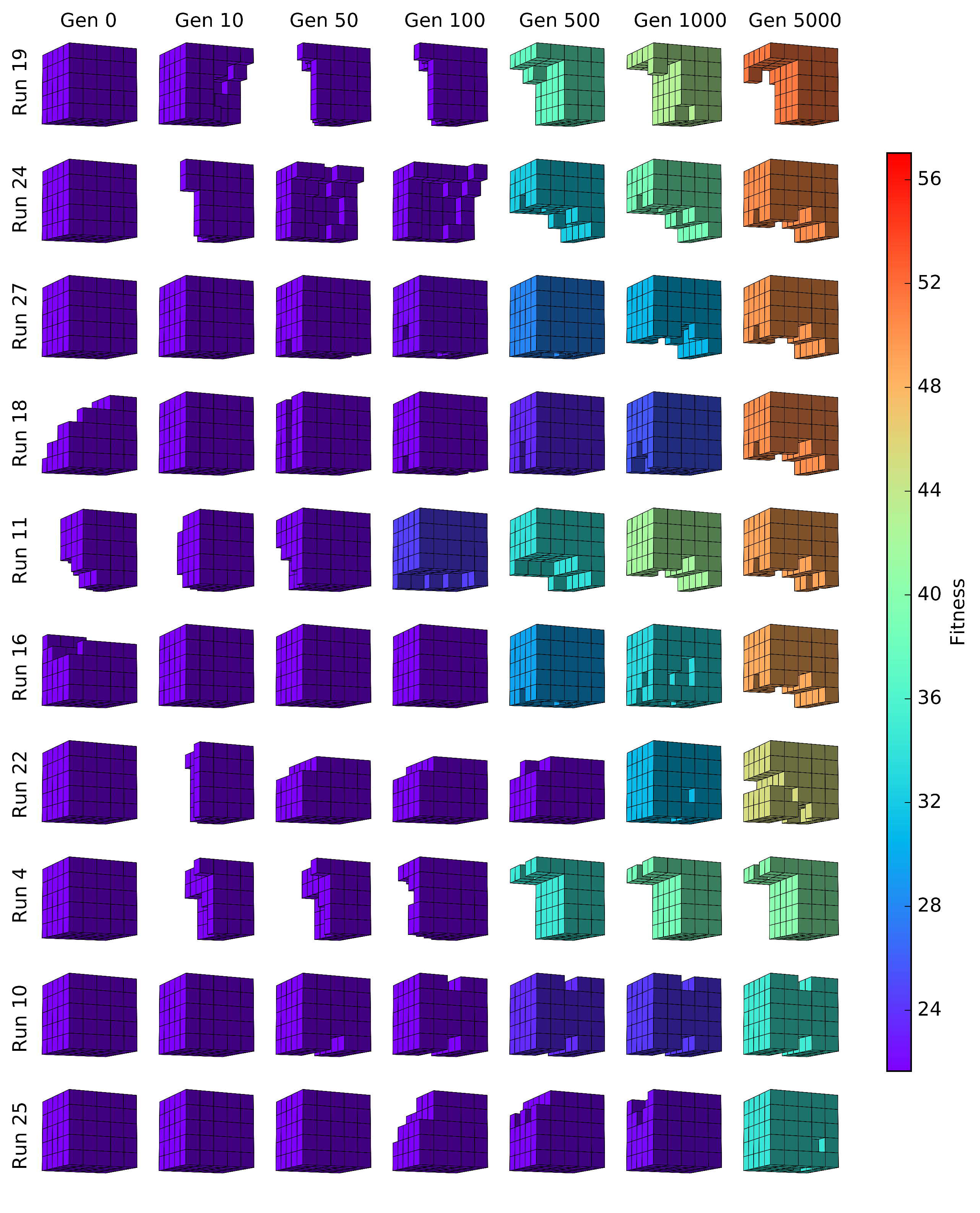}
\caption{The progression of morphologies over evolutionary time with morphological innovation protection.  Rows represent the top 10 (out of 30) performing runs at generation 5000, while columns represent snapshots of the morphology at various points during the optimization process.  Note how some of the runs converge upon the same morphology (a front and back \textquote{legged} robot), despite starting from varying initial conditions.  The color of the morphologies represent their fitness values.  
%Note the progression from cooler to warmer colors over optimization time, as fitness values increase.
}
\label{morphologyOverTime_protection}
\end{figure}

Visual inspection of actual morphologies over evolutionary time further supports the proposed method's improved optimization efficiency and ability to escape local optima.  Fig.~\ref{morphologyOverTime_protection} shows the current best individual at various points over evolutionary time for each of the top 10 runs in the treatment with morphological innovation protection.  Note how the fitness values of the robots increase over time (from left to right, and indicated by the color of each robot).  Also note how the final morphology of some robots (e.g. runs 24, 27, 18, 11, and 16) result in identical morphologies, despite starting from a range of starting morphologies and not finding this convergent morphology until hundreds or thousands of generations into the optimization process.

\begin{figure}[t!]
\includegraphics[width=\linewidth]{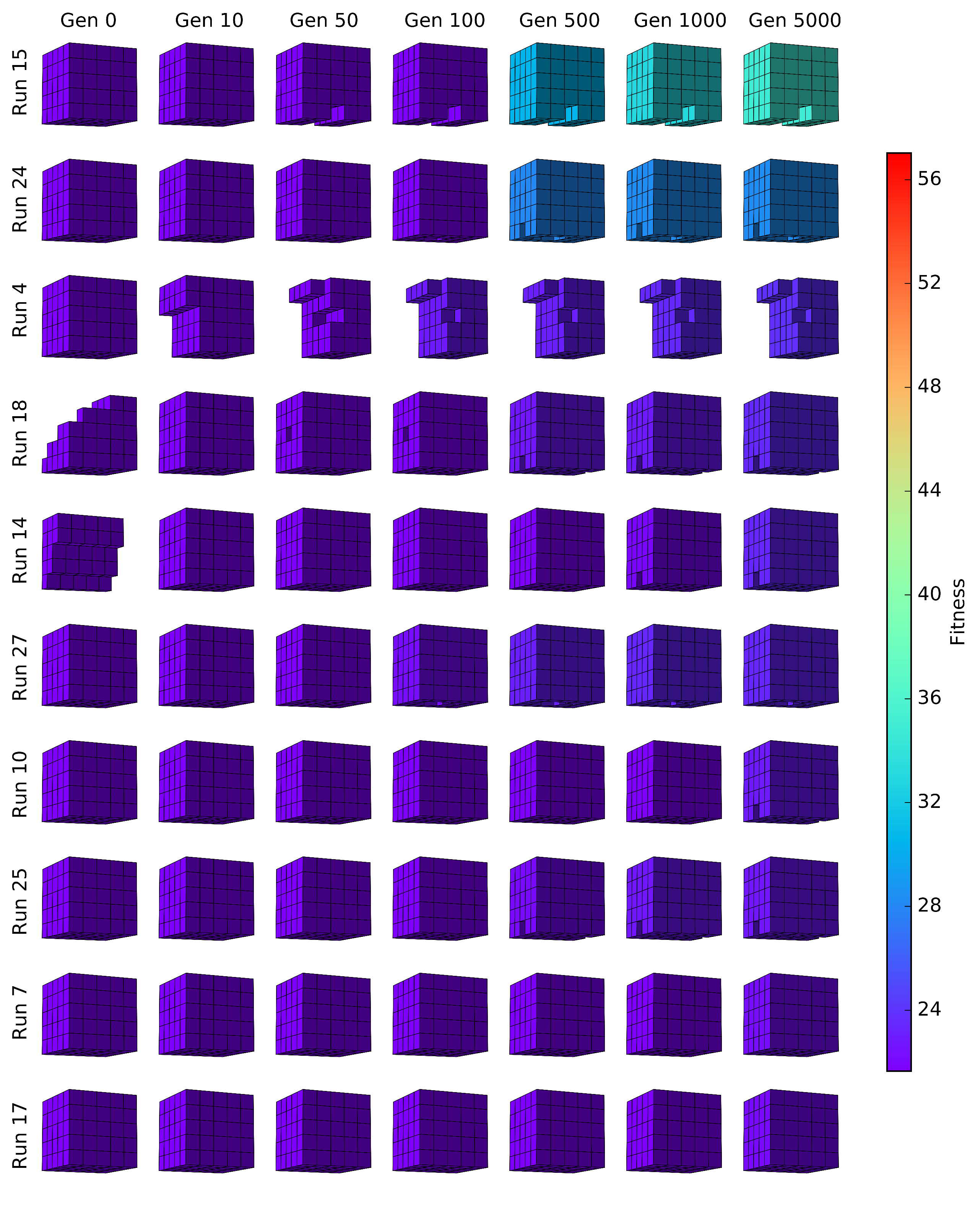}
\caption{The progression of morphologies over evolutionary time for the top 10 performing runs with no innovation protection.  The color legend and column snapshots times are identical to Fig.~\ref{morphologyOverTime_protection}.  
%Rows represent the top 10 performing runs in this treatment, while columns represents the progression of optimization over evolutionary time. The color of the morphologies represent their fitness values, and share the same color mapping from fitness to color as Fig~\ref{morphologyOverTime_protection}.  Note the lack of improvement in fitness (color) or change in morphologies over evolutionary time, as the final morphologies found at generation 5000 tend to resemble the state of their run as early as generation 50 -- just 1\% of the way into optimization time.
}
\label{morphologyOverTime_noProtection}
\end{figure}

In contrast to the sustained turnover of morphologies shown above, Fig.~\ref{morphologyOverTime_noProtection} shows the snapshots of the 10 best runs in the treatment without innovation protection.  Notice how the colors of the robots tend to show little change over the evolutionary process, mirroring the stagnation shown in Fig.~\ref{fitnessOverTime}.  While convergence of the final morphologies is present here as well, the gross morphologies found here (variants of the a full cube with no appendages) are found early on in the search -- and often provided in the set of random initial morphologies.
% (an artifact of this genetic encoding's tendency to start with simple shapes and complexify them over time).
%, or found early on in search (by generation 10).  
In this treatment, gross morphological changes tend to be absent after generation 50 (just 1\% into the full 5000 generations).

\begin{figure}[t!]
\includegraphics[width=\linewidth]{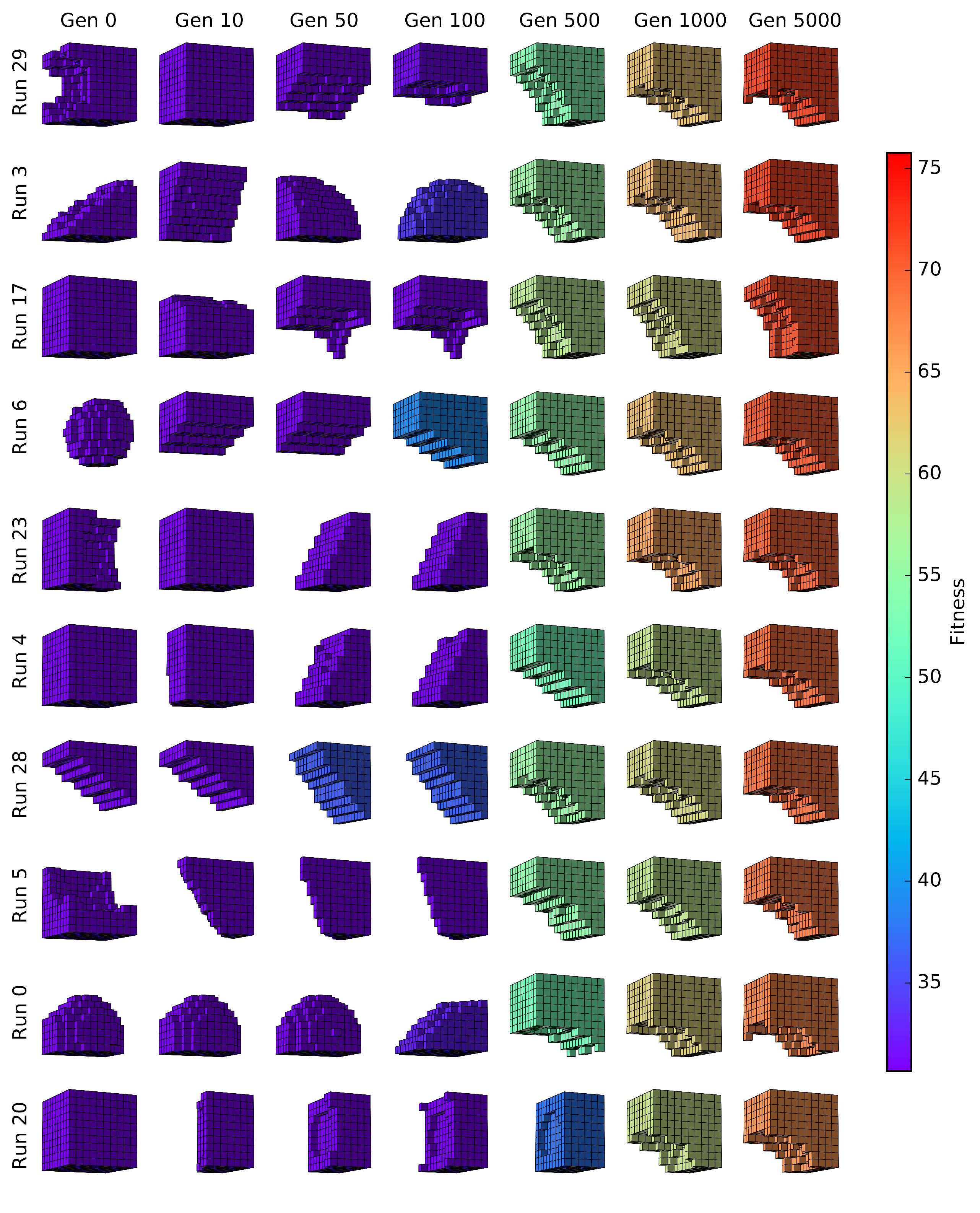}
\caption{The progression of morphologies over evolutionary time in the setting with high-resolution morphologies and evolution with morphological innovation protection with a 20\% threshold.  The rows represent the top 10 (of 30) independent trials, while the columns represent the progression over evolutionary time.  Color represent the fitness values of the robot (their locomotion speed), with warmer values depicting more fit individuals.  Note the convergence of all 10 of these runs to the same final morphology.
% at the end of optimization time. 
%(and many of the these runs find this gross morphology by generation 500).  This convergence occurs despite a variety of different initial conditions across these trials.
}
\label{morphologiesOverTime_phaseOffset_size10_withProtection_0.2Thresh}
\end{figure}

The differences between Figs.~\ref{morphologyOverTime_protection} and \ref{morphologyOverTime_noProtection} suggests that the traditional method without morphological innovation protection tends to converge prematurely to morphologies early in the evolutionary search, while morphological innovation protection may allow search to escape these local optima and converge to \textquote{more global} optima.  
%Since the common final morphology found in runs 24, 27, 18, 11, and 16 of Fig.~\ref{morphologyOverTime_protection} is not the best found in that treatment (as it is outperformed by run 19), it is obviously not the global optima.  But the fact that a diversity of random initial conditions converged to this final morphology (and that the converged upon morphology is more fit than those initial conditions) suggests that morphological innovation protection is better able to search over a larger basin than treatments without it (i.e. are less sensitive to initial conditions and can better escape local optima).
%
The convergence of morphologies across varied initial conditions is even further pronounced in Fig.~\ref{morphologiesOverTime_phaseOffset_size10_withProtection_0.2Thresh}.

%\edit{Continual improvement justified by increase in age zero fitness over time}

%\subsection{Morphological Innovation Protection Can Lead To Increased Convergence Across Trials}

%
%\begin{figure}[t!]
%% \includegraphics[width=0.95\linewidth]{oldImages/softbotVisualization_2015-11-02_voxelResetThresh_0_001_restart_allInOne.png}
%%\includegraphics[width=0.95\linewidth]{Figure4.png}
%% \includegraphics[width=0.95\linewidth]{Fig4.eps}
%\includegraphics[width=\linewidth]{FigS8.jpg}
%\caption{The progression of the best performing robot morphology at various stages over optimization time with open-loop phase-offset controllers and high-resolution morphologies and evolution with morphological innovation protection without a threshold.  Each row represent one of the top 10 performing out of 30 independent trials, while each column shows that trial's best performing morphology at a given time, colored according to their fitness.  The color mapping here is consistent with that in Fig.~\ref{morphologiesOverTime_phaseOffset_size10_withProtection_0.2Thresh}.  This treatment does not show the same convergence as Fig.~\ref{morphologiesOverTime_phaseOffset_size10_withProtection_0.2Thresh}, but does demonstrate fitness improvements over time, and some runs are able to find the morphology which is converged upon in the 20\% threshold case.}
%\label{morphologiesOverTime_phaseOffset_size10_withProtection_noThresh}
%\end{figure}

\begin{figure}[t!]
\includegraphics[width=\linewidth]{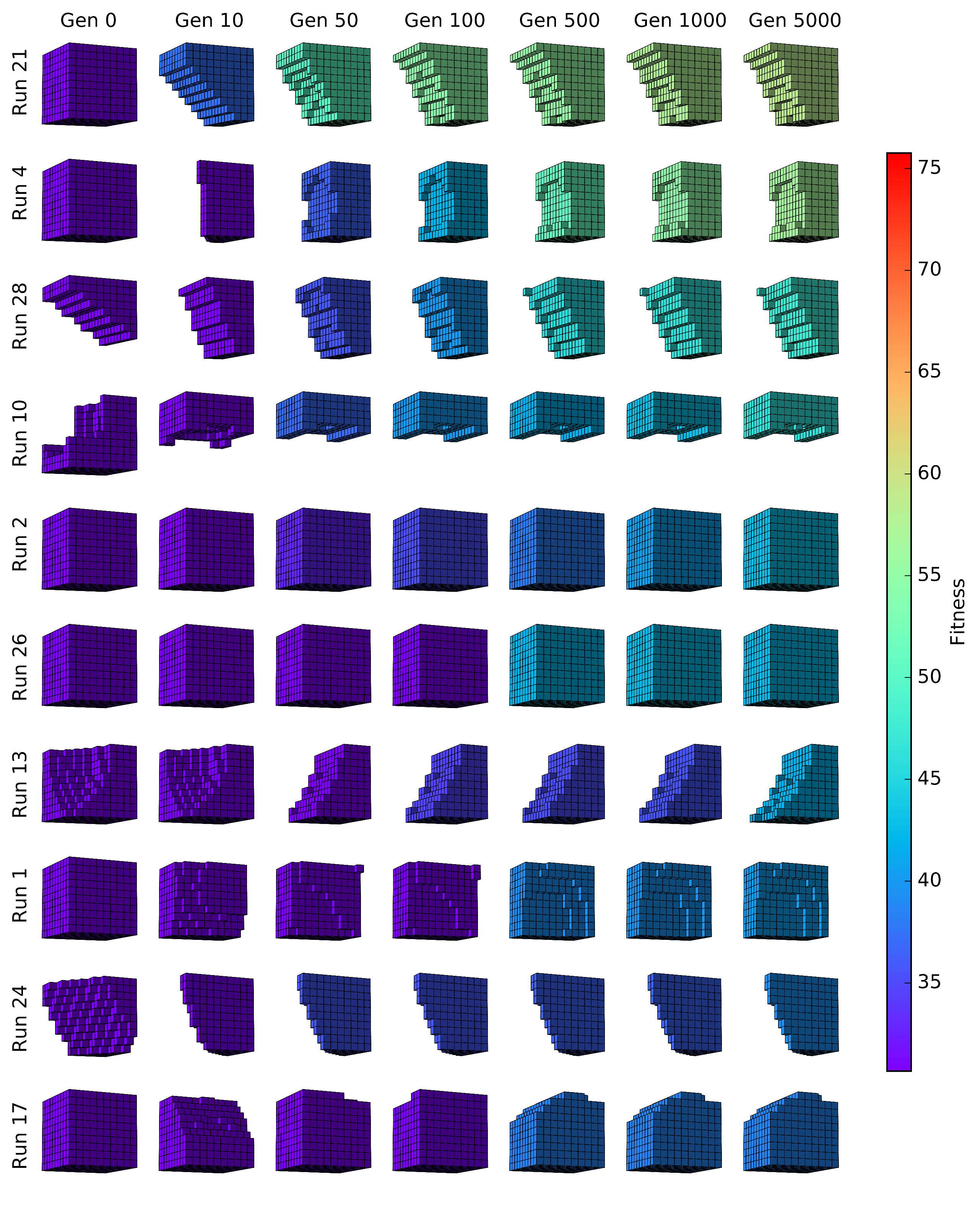}
\caption{
The progression of high-resolution morphologies over evolutionary time for the top 10 performing runs with no innovation protection.  The color legend and column snapshots times are identical to Fig.~\ref{morphologiesOverTime_phaseOffset_size10_withProtection_0.2Thresh}.
%
%The progression of morphologies over evolutionary time in the setting with high-resolution morphologies and evolution with no innovation protection.  The rows represent the top 10 independent trials, while the columns represent the progression over evolutionary time.  Color represent the fitness values of each robot, and is consistent with that in Figs.~\ref{morphologiesOverTime_phaseOffset_size10_withProtection_0.2Thresh}.
% and~\ref{morphologiesOverTime_phaseOffset_size10_withProtection_noThresh}.  
%Note the lack of high fitness values and lack of morphological change, with most runs locking into a gross morphology by generation 50.
}
\label{morphologiesOverTime_phaseOffset_size10_noProtection}
\end{figure}

\subsection{The Potential for Convergence Across Initial Conditions}

%The results above demonstrate the effectiveness of the proposed algorithm on one specific instance of the optimization of embodied machines.  However, the wide application of this algorithm also requires the demonstration of its effectiveness as the complexity of these machines scales up.  

To explore the question of scale, we apply morphological innovation protection to the evolution of robots with higher resolution morphologies -- up to $10^3 = 1000$ voxels rather than the lower resolution morphologies ($5^3 = 125$ voxels) employed in the previous experiments. 
% and also those with closed-loop neural network controllers.  These both represent more complex instantiations than the lower resolution morphologies ($5^3 = 125$ voxels) and open-loop phase-offset controllers employed in the previous experiments.% (see Methods for more details).  

%In this particular robot morphology and encoding, t
The increased number of voxel cells that make up each robot allows for greater expressiveness and finer details in its morphology.  However this also presents a challenge for the above algorithm.  As the number of cells increases, the effect of changing a single voxel (the minimum morphological variation) is reduced.  
%In the extreme where changes to the morphology do not create large-scale changes to the functionality of the morphology.  
In the extreme, the concept of readapting controllers since the last \textquote{morphological change} is less straightforward -- as increasingly small changes can modify minor details of the morphology without affecting its overall function.  

To help address the problem of non-functional morphology changes, we introduce a parameter to represent the minimum percentage of voxels that must be varied in order to qualify as a \textquote{gross morphological change}.  It is important to note that this parameter is specific to the voxel-based soft robot implementation employed in this work -- and thus the optimal setting of this parameter is not of great importance for its generalization outside of this soft robot encoding.  But the general concept of a threshold for the minimum morphological change is a universal concept that could be applied to any robot instantiation, as necessary.

In the case of robots with higher-resolution morphologies, we find that resetting the \textquote{morphological age} of an individual only after a mutation that changes more than 20\% of their voxels produces optimal results.  The value was found via a parameter sweep (of minimum age-resets of 0\%, 5\%, 10\%, 20\%, 30\%, 40\%, and 50\% of voxels changed).
%, shown in Fig.~\ref{thresholdParameterSweep}.  
We also investigated a minimum change threshold for controller innovation protection and found that its benefit falls as the threshold increases (showing optimal performance with no threshold), so we ignore the threshold for minimum controller change here.  
%Similarly to above, we find the optimal threshold value with a parameter sweep, this time resulting in an optimal value of 20\%.  That the threshold in this scenario differs from the optimal 40\% value in the case above is interesting to note, as it suggests that the optimal threshold value for minimum morphological changes may be dependent on both the implementation of the morphology and that of the controller.  The differences in threshold values for different controllers makes intuitive sense, as some controllers (like the open-loop distributed phase offset oscillating controller) may be more robust to changes in morphologies than other (like the closed-loop neural network controllers), and thus better able to generalize across closely related morphologies.  

%The fitness-over-time curves of optimization in the case of this high-resolution morphology are depicted in Fig.~\ref{fitnessOverTimeHiResPhaseOffset}.  
In this setting, basic morphological innovation protection (with no threshold) travels significantly farther (60.728 voxels) than both the case with no protection (32.575 voxels, $ p = 1.208*10^{-4} $) and the case with controller innovation protection (35.375 voxels, $ p = 0.018 $).  
%While visually there appears to be an advantage for employing the threshold on morphological changes, the difference in final fitness values at generation 5000 is not significant at the $p=0.05$ level (as the treatment with no morphological change threshold travels 43.340 voxels,$p=0.063$).  
While trials employing the threshold on morphological changes did achieve higher average fitness than the case of morphological innovation protection without a threshold (where robots traveled 43.340 voxels on average), this difference was not significant at the $p=0.05$ level ($p=0.063$).  

As a side note, in other more difficult trials (where the robots need to optimize a closed-loop neural network controller -- not depicted visually due to space constraints), trials with the morphological change threshold significantly outperform all other treatments (as they travel 31.313 voxels -- while trials with no protection, controller protection, and morphological innovation protection with no threshold travel 13.941, 15.356, and 17.224 voxels, respectively; all $p \leq  7.287*10^{-8} $).
%, while the morphological innovation protection with no threshold travels 17.224 voxels and outperforms the case with no protection (13.941 voxels, $p=0.019$), but fails to outperform the trails with controller innovation protection (15.356 voxels, $p=0.584$).  Thus, this morphological change threshold may be of importance in some robotic environments or tasks.  

%The 20\% threshold value again shows a significant improvement over the treatment with no morphological innovation protection ($ p = 1.208*10^{-5} $), shown in Fig.~\ref{fitnessOverTimeHiResPhaseOffset}.  
%Though interestingly, the morphological change threshold does not provide a significant fitness improvement over morphological innovation protection with no threshold in this setting ($p=0.156$).  Also note, in Fig.~\ref{fitnessOverTimeHiResPhaseOffset}, how the fitness curves for this thresholded value appear to flatten out later in optimization, as fitness gains slow down.  The explanation for this may come in Fig.~\ref{morphologiesOverTime_phaseOffset_size10_withProtection_0.2Thresh}, where most of the runs depicted have converged to the same gross morphology by generation 500, and only make fitness improvements from controller changes thereafter.
%{\color{red} what about no protection significance?}

Visualization of these soft robot morphologies over time in the trials with a 20\% morphological change threshold is depicted in Fig.~\ref{morphologiesOverTime_phaseOffset_size10_withProtection_0.2Thresh} and compared to the case with no protection in Fig.~\ref{morphologiesOverTime_phaseOffset_size10_noProtection}.
% demonstrates extreme convergence of morphologies across varying initial conditions. 
The ability of evolution to converge to the same high-performing morphology across many independent trials, despite starting from  different initial conditions suggests that (in this particular soft robot implementation) the inclusion of thresholded morphological innovation protection is able to escape the local optima around these starting conditions and find \textquote{more global} optima in this search space.  
% removed for space
%The lack of this convergence in Fig.~\ref{softbotVisNeuralNetProectThresh04} (the case of closed-loop neural network controllers) may suggest a differently structured search space
% -- potentially including fewer optima with wide basins -- 
%or simply the inefficiency of the algorithm to find such optima in the allotted time with a more complex controller to optimize.  

%This convergence to the perhaps (more) optimal morphology in the open-loop controller setting is visually less pronounced without the threshold for minimum morphological change (Fig.~\ref{morphologiesOverTime_phaseOffset_size10_withProtection_noThresh}) -- despite the fact that the fitness values at the end of optimization do not show significant differences.  Finally, in 
In
the case without any protection, search stagnates quickly and again appears unable to escape the local optima near its initial conditions (Fig.~\ref{morphologiesOverTime_phaseOffset_size10_noProtection}).

Interestingly, the low resolution soft robot implementation (in Figs.~\ref{morphologyOverTime_protection} and \ref{morphologyOverTime_noProtection}) does not benefit from the inclusion of a threshold.  
As presumably the more discrete construction of these robots mean that all mutations are large enough to create a meaningful \textquote{morphological change}.

\section{Discussion}
% \begin{itemize}
% % \item{todo}
% \item{end goal is to create generalists -- which would mean allowing stimuli-dependent development of both morphology and controller}
% \end{itemize}

The above results demonstrate a new method for entire robot evolution (i.e. both brain and body-plan) that is more scalable in terms of continued optimization for longer periods of time and better resulting fitness than the traditional evolutionary methods.
% (without innovation protection).  
This method for \textquote{morphological innovation protection} helps prevent premature convergence to the many local optima which appear to be present in the rugged search space of robot morphologies and controllers~\cite{cheney15difficulty}.  

The hypothesis from~\cite{cheney15difficulty} that the fragile co-optimization of brain and body plan caused by specialization of one sub-component to the other is consistent with the findings above.  This work also reveals that there is an asymmetry between the brain and body plan: protecting innovations to the morphology leads to more effective optimization that protecting innovations to the controller.  

The benefits of the temporarily reduced selection pressure provided by morphological innovation protection suggests that the long-term potential and immediate fitness impact of a morphological mutation are not always correlated.   Thus, we require a form of diversity maintenance to help evolution to rate proposed solutions based on their long-term potential rather than on immediate fitness impact.  As was shown here, using morphological innovation protection for this purpose can help to reduce premature convergence in the search space and stagnation at suboptimal values.  

%{ \color{red} [TODO: relation to other diversity protection mechanisms (like speciation), and relation to parallel hill-climbers] }
%We should point out that many other diversity maintenance methods already exist (such as fitness sharing~\cite{sareni1998fitness}, novelty~\cite{lehman2008exploiting}, speciation~\cite{stanley2002evolving}, or binning/niching~\cite{mouret2015illuminating}).  Yet we believe the proposed method here to be unique from these other forms of diversity maintenance, as we reduce selection pressure only for individuals with new morphologies (and not those with new controllers) to allow them to adapt.  It could be argued that this takes place also by parallel (or random restart) hillclimbers~\cite{mahfoud1995comparison} that are instantiated with a diverse set of morphologies, or certain forms of speciation -- but such methods would not allow for the direct competition between evolving individuals, based on their current level of controller-to-morphology specialization, and thus may not perform as efficiently (though future work is needed to compare the proposed method against all such competitors systematically).  

We believe this to be the first example of a design automation algorithm for robotics that considers the interdependence of neural controllers and body plans arising from the psychological theory of embodied cognition~\cite{pfeifer2006body} and uses this intuition to propose a method to escape local optima in the fitness landscape.
% of embodied machines.

Despite the significant improvement to our ability to simultaneously optimize the brain and body plan of embodied robots, there is much work still to be done.  Firstly, the proposed method was only applied to one class of robot. This class may actually represent the simplest form of brain-body co-optimization because the distributed sensing, actuation, and information processing that the cellular soft robot paradigm was designed to possess helps to blur the line between physical interactions of the morphology with the environment and information processing of a controller~\cite{cheney2013unshackling,cheney2014evolved}.  

In the case of centralized controllers and robots composed of rigid components, topological (rather than parametric) changes to the cognitive architecture would be required for control readaptation if morphological mutations add or remove physical components (e.g. limbs).  
% Following our hypothesis on the specialization of morphologies and controllers, such drastic morphological change would necessitate controller readaptation in order to apply new low-level behaviors to each new morphological variant.  
Future work should explore the effect of morphological innovation protection in such a paradigm, where there is the potential for morphological changes to more drastically change the function of the robot -- and thus for readaptation to those morphologies to play an even more critical role in optimization.  

In these experiments, the genotype encoding of the soft robots was modularized such that one part of the genome dictated the shape and material properties of the robot and a separate part encoded the actuations in the form of volumetric deformations of the voxels during behavior.  In these robots (and those with even more complex phenotypes), one can envision different splits as to what constitutes morphology and control -- and what qualifies as a \textquote{morphological change} in this proposed algorithm. It would be of interest to investigate the effect of various splitting points of this dichotomy on the efficacy of morphological innovation protection.  
%{\color{red} \st{At the extreme one can imagine a split which simply reverses the names of the two modules in this experiment, and thus the idea of morphological innovation protection would no longer hold as the morphology module would contain what was declared in this work to be the controller (and was shown to be less effective to protect).  This split was chosen logically, but also somewhat naively here, and it is likely be of importance.}}  

Furthermore, the very idea of a distinct split between body and brain is a false dichotomy, as information processing and physical processes happen throughout the agent~\cite{baluska2016having}.  Rather than various mutation operators which affect only the %genotypes of the 
\textquote{brain} or the \textquote{body-plan}
of a robot,
we desire agents that grown and develop as the result of rich crosstalk and feedback loops between the processes responsible for creating both of these components (creating complex adaptive systems like we see in biological organisms).  
How this proposed method relates to, and informs, biological analogs
not entirely clear, but some rough analogies can be drawn.  For example, the observation that brains learn and adapt at a much faster rate than bodies grow fits into this paradigm, as does the even slower change of gross morphological features over evolutionary time.  In this way, the readaptation of controller strategies for varying body plans is built into the neuroplasticity of the brain.  %The speed at which neurogenesis and morphogenesis occur is certainly constrained by the energetic resources of each process, but herein lies a potential benefit for the automated optimization of behavior and form in machines.  
%The basic idea of \textquote{protection} is simply a diversity maintenance measure resulting in temporarily reduced selection pressure on specific individuals within an evolving population.
%{\color{red} [can't have diversity maintenance acting on an individual, only a population.rephrase.]}  

One could also imagine periods of evolutionary time when an entire species is under relatively little morphological selection pressure before an environmental shock suddenly reapplies that pressure.  Also possible are periods of a single individual's lifetime when selection pressure varies: For example, human infants may not be selected highly for their locomotion speed, as their parents tend to physically carry and protect them early in life.
We do not believe that the methods introduced here are restricted to this particular domain.  The above algorithm is simple to implement (requiring only: an age counter, a check for variations in brain and/or body for each mutation, and --optionally -- a criterion for the minimal gross morphological change), and thus we believe it will be widely applicable.  Future work will test this supposition.  

% {\color{red} [talk about nested optimization of neural networks, e.g pathnet, EWC]}
Due to the recent interest in co-optimization of neural network topology and weights~\cite{fernando2017pathnet, miikkulainen2017evolving}, we should also note that this work -- an agent's controller embodied within its morphology -- is closely related to that of neural network's weights embodied within its topology.  Future work will show whether the method proposed here will show similar optimization gains in the design of neural network topologies as well.  

\section{Conclusion}
%{\color{red} [The conclusion goes here.]}
We demonstrate an example of a robot design automation algorithm that considers the interdependence of neural controllers and body plans (due to the theory of embodied cognition) on the optimization process.  We use this intuition to temporarily reduce selection pressure on newly mutated robot morphologies, thus allowing the agents to readapt their controllers and better escape local optima in the fitness landscape.  We have shown that this technique -- deemed \textquote{morphological innovation protection} -- produces evolutionary optimization which delays premature convergence and stagnation, and results in more efficient evolved robots.  We showcase the ability of this technique to escape local optima in the search space by demonstrating the convergence to a similar morphology across many independent trials from randomly initial conditions.  While we hope that this technique will be surpassed in the future by a developmental process with feedback loops between the body and brain, we propose the above algorithm as a short term improvement over the current techniques for the co-optimization of morphology and control in virtual creatures.

% if have a single appendix:
%\appendix[Proof of the Zonklar Equations]
% or
%\appendix  % for no appendix heading
% do not use \section anymore after \appendix, only \section
% is possibly needed

% use appendices with more than one appendix
% then use \section to start each appendix
% you must declare a \section before using any
% \subsection or using \label (\appendices by itself
% starts a section numbered zero.)
%

%\appendices
%\section{Proof of the First Zonklar Equation}
%Appendix one text goes here.
%
%% you can choose not to have a title for an appendix
%% if you want by leaving the argument blank
%\section{}
%Appendix two text goes here.
%
%%
%\section{Supporting Information}

%\renewcommand{\figurename}{SI Figure}
%\renewcommand{\thefigure}{S\arabic{figure}}

\setcounter{table}{0}
\renewcommand{\thetable}{S\arabic{table}}%
\setcounter{figure}{0}
\renewcommand{\thefigure}{S\arabic{figure}}%

% use section for acknowledgment
\section{Acknowledgment}

We thank NASA Space Technology Research Fellowship \#NNX13AL37H and Army Research Office Contract \#W911NF1610304 for support, Steve Strogatz for advice, Kathryn Miller for copy editing, and the UVM Morphology, Evolution \& Cognition Lab for feedback.%, including Anton Bernatskiy. 

% Can use something like this to put references on a page
% by themselves when using endfloat and the captionsoff option.
\ifCLASSOPTIONcaptionsoff
  \newpage
\fi

% trigger a \newpage just before the given reference
% number - used to balance the columns on the last page
% adjust value as needed - may need to be readjusted if
% the document is modified later
%\IEEEtriggeratref{8}
% The "triggered" command can be changed if desired:
%\IEEEtriggercmd{\enlargethispage{-5in}}

% references section

% can use a bibliography generated by BibTeX as a .bbl file
% BibTeX documentation can be easily obtained at:
% http://mirror.ctan.org/biblio/bibtex/contrib/doc/
% The IEEEtran BibTeX style support page is at:
% http://www.michaelshell.org/tex/ieeetran/bibtex/
%\bibliographystyle{IEEEtran}
% argument is your BibTeX string definitions and bibliography database(s)
%\bibliography{IEEEabrv,../bib/paper}
%
% <OR> manually copy in the resultant .bbl file
% set second argument of \begin to the number of references
% (used to reserve space for the reference number labels box)
%\begin{thebibliography}{1}
%
%\bibitem{IEEEhowto:kopka}
%H.~Kopka and P.~W. Daly, \emph{A Guide to \LaTeX}, 3rd~ed.\hskip 1em plus
%  0.5em minus 0.4em\relax Harlow, England: Addison-Wesley, 1999.
%
%\end{thebibliography}
\bibliographystyle{abbrv}
\bibliography{softbots.bib}

\end{document}